\documentclass[epsf, twocolumn]{IEEEtran}
\usepackage{graphicx}

\def\be{\begin{eqnarray}}
\def\en{\end{eqnarray}}
\def\bee{\begin{eqnarray*}}
\def\enn{\end{eqnarray*}}

\begin{document}

\title{Rational Radial Distortion Models with Analytical Undistortion Formulae}
\author{Lili Ma, {\it Student Member, IEEE}, YangQuan Chen and Kevin L. Moore, {\it Senior Members, IEEE}\\ Center for Self-Organizing and Intelligent Systems (CSOIS),\\Dept. of Electrical and Computer Engineering, 4160 Old Main Hill,\\ Utah State University (USU), Logan, UT 84322-4160, USA.\\
Emails: \texttt{lilima@cc.usu.edu, \{yqchen, moorek\}@ece.usu.edu}}
\maketitle{}

\begin{abstract}
The common approach to radial distortion is by the means of polynomial approximation, which introduces distortion-specific parameters into the camera model and requires estimation of these distortion parameters. The task of estimating radial distortion is to find a radial distortion model that allows easy undistortion as well as satisfactory accuracy. This paper presents a new class of rational radial distortion models with easy analytical undistortion formulae. Experimental results are presented to show that with this class of rational radial distortion models, satisfactory and comparable accuracy is achieved.
\end{abstract}

\thispagestyle{empty}
\section{Introduction}

\subsection{Camera Calibration}
To address the problem of radial distortion, the problem of camera calibration needs to be addressed first, since the radial distortion is one step in the camera calibration procedures. Depending on what kind of calibration object used, there are mainly two categories of calibration methods: photogrammetric calibration and self-calibration. Photogrammetric calibration refers to those methods that observe a calibration object whose geometry in 3-D space is known with a very good precision \cite{Emanuele98introductorycomputervision,zhang99calibrationinpaper,OlivierF01Straight}. Self-calibration does not need any calibration object. It only requires point matches or correspondences from image sequence. In \cite{Faugeras92self-calibration}, it is shown that it is possible to calibrate a camera just by pointing it to the environment, selecting points of interest and then tracking them in the image as the camera moves. The obvious advantage of the self-calibration method is that it is not necessary to know the camera motion and it is easy to set up. The disadvantage is that it is usually considered unreliable \cite{Bougnoux98criticismself-calibration}. A four step calibration procedure is proposed in \cite{Heikkil97fourstepcameracalibration} where the calibration is performed with a known 3-D target. The four steps in \cite{Heikkil97fourstepcameracalibration} are: linear parameter estimation, nonlinear optimization, correction using circle/ellipse, and image correction. But for a simple start, linear parameter estimation and nonlinear optimization are enough. In \cite{STURM99planebasedcalibrationsigularities}, a plane-based calibration method is described where the calibration is performed by first determining the absolute conic ${\bf B} = {\bf A}^{-T} {\bf A}^{-1}$, where $\bf A$ is a matrix formed by the camera's intrinsic parameters. In \cite{STURM99planebasedcalibrationsigularities}, the parameter $\gamma$ (a parameter describing the skewness of the two image axes) is assumed to be zero and it is observed that only the relative orientations of planes and camera are of importance in avoiding singularities because the planes that are parallel to each other provide exactly the same information. The camera calibration method in \cite{zhang99calibrationinpaper} focuses on the desktop vision system and lies between the photogrammetric calibration and the self-calibration, because 2-D metric information is used rather than 3-D. The key feature of the calibration method in \cite{zhang99calibrationinpaper} is that the absolute conic $\bf B$ is used to estimate the intrinsic parameters and the parameter $\gamma$ can be considered. The proposed technique in \cite{zhang99calibrationinpaper} only requires the camera to observe a planar pattern at a few (at least 3, if both the intrinsic and the extrinsic parameters are to be estimated uniquely) different orientations. Either the camera or the calibration object can be moved by hand as long as they cause no singularity problem and the motion of the calibration object or camera itself needs not to be known in advance.

After estimation of camera parameters, a perspective projection matrix $\bf M$ can directly link a point in the 3-D world reference frame to its projection (undistorted) in the image plane. That is
\begin{eqnarray}
\label{eqn: projection matrix}
\lambda \left [\matrix{u \cr v \cr 1} \right ] 
&=& {\bf M} \left [\matrix{X^w \cr Y^w \cr Z^w \cr 1} \right ] 
  = {\bf A} \, [{\bf R} \mid {\bf t}] \left [\matrix{X^w \cr Y^w \cr Z^w \cr 1} \right ] 
  = {\bf A} \left [\matrix{X^c \cr Y^c \cr Z^c} \right ], \nonumber\\
\left [\matrix{u \cr v \cr 1} \right ]
&=& {\bf A} \left [\matrix{\frac{X^c}{Z^c} \cr \frac{Y^c}{Z^c} \cr 1} \right ]
  = \left [ \matrix {
\alpha & \gamma &u_0 \cr
0 & \beta & v_0 \cr
0 & 0 & 1
} \right ] \left [\matrix{x \cr y \cr 1} \right ].
\end{eqnarray}
The matrix $\bf A$ fully depends on the 5 intrinsic parameters with their detail descriptions in Table \ref{table: variables used}, where some other variables used throughout this paper are also listed.

The calibration method used in this work first estimates the projection matrix and then uses the absolute conic to estimate the intrinsic parameters \cite{zhang99calibrationinpaper}. The detail procedures are summarized below:
\begin{itemize}
\item{Linear Parameter Estimation,}
    \begin{itemize}
    \item Estimation of Intrinsic Parameters;
    \item Estimation of Extrinsic Parameters;
    \item Estimation of Distortion Coefficients;
    \end{itemize}
\item Nonlinear Optimization.
\end{itemize}

{\footnotesize
\begin{table}[htb]
\centering
\caption{List of Variables}
\label{table: variables used}
\vspace{-2 mm}
{\small
{\begin {tabular}{|c|l|}\hline
{\bf Variable} & {\bf Description} \\[1ex]\hline
$P^w = [X^w, Y^w, Z^w]^T$    & 3-D point in the world frame\\[1ex]\hline
$P^c = [X^c, Y^c, Z^c]^T$    & 3-D point in the camera frame \\[1ex]\hline
$[{\bf R} \mid {\bf t}]$     & $P^c = {\bf R} P^w + \bf t$ \\[1ex]\hline
$(u_d, \, v_d)$              & Distorted image points in pixel\\[1ex]\hline
$(u, \, v)$                  & Undistorted image points in pixel\\[1ex]\hline
$(x_d, \, y_d)$              & $[x_d, y_d, 1]^T = {\bf A}^{-1}[u_d, v_d, 1]^T$\\[1ex]\hline
$(x, \, y)$                  & $[x, y, 1]^T = {\bf A}^{-1}[u, v, 1]^T$\\[1ex]\hline
$(X_u,\, Y_u)$               & $\left[ \matrix{X_u \cr Y_u}\right] = f \, \left[ \matrix{\frac{X^c}{Z^c} \cr \frac{Y^c}{Z^c} }\right] = f \,  \left[ \matrix{x \cr y}\right]$ \\[1ex]\hline
$r$                          & $r^2 = x^2 + y^2$ \\[1ex]\hline
$f$                          & Focal length \\[1ex]\hline
$(\alpha, \beta, \gamma, u_0, v_0)$   & 5 intrinsic parameters \\[1ex]\hline
${\bf k}$                    & Distortion coefficients \\[1ex]\hline
$J$                         & Objective function \\[1ex]\hline
${\bf A}$                   & Camera intrinsic matrix \\[1ex]\hline
\end {tabular}}}   
\end{table}}

\subsection{Radial Distortion}
Virtually all imaging devices introduce certain amount of nonlinear distortion, where the radial distortion is the most severe part \cite {OlivierF01Straight,tsai87AVersatile}. Radial distortion causes an inward or outward displacement of a given image point from its ideal location. The negative radial displacement of the image points is referred to as the barrel distortion, while the positive radial displacement is referred to as the pincushion distortion \cite{Juyang92distortionmodel}. 

The removal or alleviation of radial distortion is commonly performed by first applying a parametric radial distortion model, estimating the distortion coefficients, and then correcting the distortion. Most of the existing works on radial distortion models can be traced back to an early study in photogrammetry \cite{Photogrammetry80} where the radial distortion is governed by the following polynomial equation \cite{zhang99calibrationinpaper,GWei94Implicit,Tsai88Techniques,Janne96Calibration}:
\begin{equation}
\label{eqn: general polynomial}
F(r) = r \, f(r) = r \, (1 + k_1 r^2 + k_2 r^4 + k_3 r^6+ \cdots),
\end{equation}
where $k_1, k_2, k_3, \ldots$ are the distortion coefficients and \cite{zhang99calibrationinpaper,GWei94Implicit}
\begin{equation}
r^2 = x^2 + y^2,\nonumber
\end{equation}
or \cite{Tsai88Techniques,Janne96Calibration}
\begin{equation}
r^2 = X_u^2 + Y_u^2 = f^2 (x^2 + y^2). \nonumber
\end{equation}
Both of the above two formulae of $r$ are in the camera frame and they are basically the same, since the resulting distortion coefficients have one-to-one relations with each other, where one set of distortion coefficients are proportional to the other set by a series of scalars $(f^2, f^4, \ldots)$. 

Until recently, the most commonly used radial distortion models are still in the polynomial form of (\ref{eqn: general polynomial}), though other models, such as the division model \cite{Andrew01Simultaneous} and the fish-eye radial distortion models (the Fish Eye Transform \cite{OlivierF01Straight} and the Field-Of-View \cite{Basu95FET}), are available in the literature.

For the polynomial radial distortion model in (\ref{eqn: general polynomial}) and its variations, the distortion is especially dominated by the first term and it has also been found that too high an order may cause numerical instability \cite{zhang99calibrationinpaper,tsai87AVersatile,GWei94Implicit}. In this paper, at most three terms of radial distortion are considered. When using two coefficients, the $f(r)$ in (\ref{eqn: general polynomial}) becomes
\be \label{eqn: polynomial 2 4}
f(r) = 1 + k_1 \, r^2 + k_2 \, r^4.
\en
The relationship between the distorted and the undistorted image points becomes \cite{zhang99calibrationinpaper}
\begin{eqnarray}
\begin{array}{c}
\label{eqn: radial distortion order 2 4}
u_d - u_0 = (u-u_0) \, (1 + k_1 \, r^2 + k_2 \, r^4), \\
v_d - v_0 = (v-v_0) \, (1 + k_1 \, r^2 + k_2 \, r^4).
\end{array}
\end{eqnarray}
The polynomial function in (\ref{eqn: polynomial 2 4}) has one main drawback, that is, the inverse of the polynomial function in (\ref{eqn: radial distortion order 2 4}) is difficult to perform analytically but can be obtained numerically via an iterative scheme. In \cite{Undistortionchapter}, for practical purpose, only one distortion coefficient $k_1$ is used. To overcome the inversion problem, another polynomial radial distortion model using also two terms is proposed as \cite{LiliISIC03Flex}
\be \label{eqn: polynomial 1 2} f(r) = 1 + k_1 \, r + k_2 \, r^2, \en
whose main appealing feature lies in its satisfactory accuracy as well as the existence of an easy analytical radial undistortion formula. The two polynomial radial distortion models in (\ref{eqn: polynomial 2 4}) and (\ref{eqn: polynomial 1 2}) act as benchmarks for evaluating the performance of the rational distortion models presented in Sec. \ref{sec: rational}. 

In this work, a new class of rational radial distortion models are proposed. To compare the performance of our new models with the existing polynomial approximation models, the calibration procedures presented in \cite{zhang99calibrationinpaper} are applied, while being aware that the usage of other calibration methods, such as the image registration method in \cite{Torn02Correcting} and the plumb-line algorithm in \cite{Brown71CloseRange}, are also feasible. 

In \cite{zhang99calibrationinpaper}, the estimation of radial distortion is done after having estimated the intrinsic and the extrinsic parameters just before the nonlinear optimization step. So, for different radial distortion models, we can reuse the estimated intrinsic and extrinsic parameters. To compare the performance of different radial distortion models, the value of optimization function $J$ is used, where the initial guess for $\bf k$ is chosen to be 0. The objective function used for nonlinear optimization is \cite{zhang99calibrationinpaper}:
\begin{equation}
\label{eqn: objective function}
J = \sum_{i=1}^N \sum_{j=1}^n \|m_{i j}-\hat m({\bf A},k_1, k_2, {\bf R}_i, {\bf t}_i, M_j) \|^2,
\end{equation}
where $\hat m({\bf A},k_1, k_2, {\bf R}_i, {\bf t}_i, M_j)$ is the projection of point $M_j$ in the $i^{th}$ image using the estimated parameters and $M_j$ is the $j^{th}$  3-D point in the world frame with $Z^w = 0$. Here, $n$ is the number of feature points in the coplanar calibration object and $N$ is the number of images taken for calibration.

The rest of the paper is organized as follows. Sec.~\ref{sec: Polynomial} describes the two polynomial radial distortion models (\ref{eqn: polynomial 2 4}) and (\ref{eqn: polynomial 1 2}) in detail, where the inverse problem of (\ref{eqn: polynomial 2 4}) and the analytical undistortion formula of (\ref{eqn: polynomial 1 2}) are also described. The new class of rational radial distortion models and the comparison with the existing polynomial models are presented in Sec.~\ref{sec: rational}. Finally, some concluding remarks are given in Sec.~\ref{sec: conclusion}.

\section{Polynomial Radial Distortion Models and Their Undistortion Functions}
\label{sec: Polynomial}
From (\ref{eqn: radial distortion order 2 4}), the radial distortion can be resulted in one of the  following two ways:

\begin{itemize}

\item{Transform from the camera frame to the image plane, then perform distortion in the image plane}
\begin{eqnarray}
{\left [\matrix {
x \cr
y}\right]} \rightarrow
{\left [\matrix {
u \cr
v}\right]} \rightarrow
{\left [\matrix {
u_d \cr
v_d}\right];}\nonumber
\end{eqnarray}

\item{Perform distortion in the camera frame, then transform to the image plane}
\begin{eqnarray}
	{\left [\matrix {
	x \cr
	y}\right]} \rightarrow
	{\left [\matrix {
	x_d \cr
	y_d}\right]} \rightarrow
	{\left [\matrix {
	u_d \cr
	v_d}\right],} \nonumber
\end{eqnarray}
where
\begin{eqnarray} 	\label{eqn: xdyd and xy} 	x_d = x \, f(r), \quad y_d = y \, f(r). \end {eqnarray}
\end{itemize}
This is because
\begin{eqnarray}
\left[ \matrix{u \cr v \cr 1}\right] = {\bf A} \left[ \matrix{x \cr y \cr 1}\right] = \left [ \matrix {
\alpha & \gamma &u_0 \cr
0 & \beta & v_0 \cr
0 & 0 & 1
} \right ] \left[ \matrix{x \cr y \cr 1}\right]. \nonumber
\end{eqnarray}
Then, (\ref{eqn: radial distortion order 2 4}) becomes
\begin{eqnarray}
u_d &=& (u-u_0) \, f(r) + u_0 \nonumber \\
    &=& \alpha \, x f(r) + \gamma \, y f(r) + u_0 \nonumber\\
    &=& \alpha \, x_d + \gamma \, y_d + u_0, \\
v_d &=& (v-v_0) \, f(r) + v_0\nonumber\\
    &=& \beta \, y_d  + v_0.\nonumber
\end{eqnarray}
Therefore, 
\begin{eqnarray}
\left[ \matrix{u_d \cr v_d \cr 1}\right] = {\bf A} \left[ \matrix{x_d \cr y_d \cr 1}\right]. \nonumber
\end{eqnarray}
Thus, the distortion performed in the image plane can be understood from a new point of view: introducing distortion in the camera frame and then transform to the image plane. The radial undistortion can be perfomed by extracting $(x,y)$ from $(x_d, y_d)$.

\subsection{Radial Undistortion of Model (\ref{eqn: polynomial 2 4})}
The following derivation shows the problem when trying to extract $(x,y)$ from $(x_d, y_d)$ using (\ref{eqn: polynomial 2 4}). According to (\ref{eqn: xdyd and xy}),
\begin{eqnarray}
\begin{array}{c}
x_d = x f(r) = x [1+k_1(x^2+y^2) + k_2(x^2+y^2)^2], \\[4pt]
y_d = y f(r) = y [1+k_1(x^2+y^2) + k_2(x^2+y^2)^2].
\end{array}
\end{eqnarray}
It is obvious that $x_d = 0$ iff $x = 0$. When $x_d \neq 0$, by letting $c = y_d/x_d = y /x$, we have $y = cx$ where $c$ is a constant. Substituting $y = cx$ into the above equation gives
\begin{eqnarray}
\label{eqn: undistortion case(1)}
x_d = x \, [ 1 + k_1 (1 + c^2 )x^2 + k_2(1 + c^2)^2 x^4] .
\end{eqnarray}
Notice that the above function is an odd function. An intuitive understanding of (\ref{eqn: undistortion case(1)}) is that the radial distortion function is to approximate the relationship between $x_d$ and $x$, which, in ideal cases, is $x_d = x$. The analytical solution of (\ref{eqn: undistortion case(1)}) is not a trivial task, which is still an open problem (of course, we can use numerical method to solve it). 

\subsection{Radial Undistortion of Model (\ref{eqn: polynomial 1 2})}
\label{sec: undistortion procedures}
For (\ref{eqn: polynomial 1 2}), in a similar manner as for (\ref{eqn: polynomial 2 4}), we have
\begin{eqnarray}
\label{eqn: distortion model 3}
x_d = x[ 1 + k_1 \sqrt{1 + c^2} \, {\tt sgn}(x) \, x + k_2(1 + c^2) \, x^2],
\end{eqnarray}
where ${\tt sgn}(x)$ gives the sign of $x$. To extract $x$ from $x_d$ in (\ref{eqn: distortion model 3}), the following algorithm can be applied:

\begin{itemize}
\item[{\rm \bf 1)}]{$x = 0$ iff $x_d = 0$,}
\item[{\rm \bf 2)}]Assuming that $x > 0$, (\ref{eqn: distortion model 3}) becomes
	\bee	x_d = x + k_1 \sqrt{1 + c^2} \, x^2 + k_2(1 + c^2) \, x^3. \enn
Using {\tt solve}, a Matlab Symbolic Toolbox function, we can get three possible solutions for the above equation 
denoted by $x_{1+}$, $x_{2+}$, and $x_{3+}$ respectively. To make the equations simple, let $y = x_d$, $p = k_1 \sqrt{1+c^2}$ and $q = k_2 (1+c^2)$. The three possible solutions for $y = x + p x^2 + qx^3$ are
{\small
\begin{eqnarray}
\label{eqn: three solutions}
x_{1+} &=& \frac{1}{6q}E_1 + \frac{2}{3} E_2 - \frac{p}{3q}, \nonumber\\[1pt]
x_{2+} &=& -\frac{1}{12q} E_1 - \frac{1}{3} E_2 - \frac{p}{3q} + \frac{\sqrt{3}}{2}(\frac{1}{6q}E_1 - \frac{2}{3} E_2) \, {\bf \it j}, \\
x_{3+} &=& -\frac{1}{12q} E_1 - \frac{1}{3} E_2 - \frac{p}{3q} - \frac{\sqrt{3}}{2}(\frac{1}{6q}E_1 - \frac{2}{3} E_2) \, {\bf \it j}, \nonumber
\end{eqnarray}}
where
{\small
\begin{eqnarray}
\label{eqn: E1 E2}
E_1 &=& \{ 36pq+108yq^2-8p^3\nonumber\\
&& +12\sqrt{3}q\sqrt{4q-p^2+18pqy+27y^2q^2-4yp^3} \}^{1/3},\\
E_2 &=& \frac{p^2-3q}{qE_1}, \quad {\bf \it j} = \sqrt{-1}.  \nonumber
\end{eqnarray}}
From the above three possible solutions, we discard those with imaginary parts not equal to zero. Then, from the remaining, discard those solutions that conflict with the assumption that $x > 0$. Finally, we get the candidate solution $x_+$ by choosing the one closest to $x_d$ if the number of remaining solutions is greater than 1.

\item[{\rm \bf 3)}] Assuming that $x < 0$, there are also three possible solutions for 
\begin{eqnarray}
\label{eqn: minus case}
x_d = x - k_1 \sqrt{1 + c^2} \, x^2 + k_2(1 + c^2) \, x^3,
\end{eqnarray}
which can be written as 
\begin{eqnarray}
\label{eqn: minus case yqp}
y = x + (-p) x^2 + qx^3. 
\end{eqnarray}
The three solutions for (\ref{eqn: minus case yqp}) can thus be calculated from (\ref{eqn: three solutions}) and (\ref{eqn: E1 E2}) by substituting $p = -p$. With a similar procedure as described in the case for $x > 0$, we will have another candidate solution $x_-$. 

\item[{\rm \bf 4)}] Choose among $x_+$ and $x_-$ for the final solution of $x$ by taking the one closest to $x_d$.
\end{itemize}

The basic idea to extract $x$ from $x_d$ in (\ref{eqn: distortion model 3}) is to choose from several candidate solutions, whose analytical formulae are known. The benefits of using this new radial distortion model are as follows:
\begin{itemize}
\item Low order fitting, better for fixed-point implementation;
\item Explicit or analytical inverse function with no numerical iterations, which is important for real-time vision applications. 
\end{itemize}

\section{Rational Radial Distortion Models}
\label{sec: rational}

\begin{table*}[htb]
\centering
\caption{Distortion Models}
\label{table: Models}
\renewcommand{\arraystretch}{1}
\vspace{-2 mm}
{\small
{\begin {tabular}{|c|l|l|}\hline
{\bf Model} $\#$ & $f(r)$ & $x_d = f(x)$  \\[1.5ex]\hline
0 & $1 + k_1 \,r^2+k_2\, r^4$ & $x \cdot (1 + k_1 \,(1+c^2)\, x^2 + k_2 \,(1+c^2)^2 \, x^4)$ \\[1.5ex]
1 & $1 + k \,r$   & $x \cdot(1 + k \sqrt{1+c^2} \, x \, sgn(x))$\\[1.5ex]
2 & $1 + k \,r^2$ & $x \cdot (1 + k\,(1+c^2)\, x^2)$\\[1.5ex]
3 & $1 + k_1 \,r+k_2\, r^2$ & $x \cdot (1 + k_1 \,\sqrt{1+c^2} \,x \,sgn(x) + k_2\,(1+c^2) \, x^2)$\\[1.8ex]
4 & $\displaystyle \frac{1}{1 + k\, r}$ & $x \cdot \displaystyle \frac{1}{1+k \,\sqrt{1+c^2}\, x \,sgn(x)}$\\[2.2ex]
5 & $\displaystyle \frac{1}{1 + k \,r^2}$ & $x \cdot \displaystyle \frac{1}{1+k \,(1+c^2)\,x^2}$\\[3ex]
6 & $\displaystyle \frac{1+k_1\, r}{1 + k_2\, r^2}$ & $x \cdot \displaystyle \frac{1+k_1 \,\sqrt{1+c^2}\, x\, sgn(x)}{1+k_2 \,(1+c^2)\,x^2}$\\[2.8ex]
7 & $\displaystyle \frac{1}{1 + k_1\, r + k_2\, r^2}$ & $x \cdot \displaystyle \frac{1}{1+k_1 \sqrt{1+c^2}\, x\, sgn(x) + k_2 \,(1+c^2)\,x^2}$\\[3.5ex]
8 & $\displaystyle \frac{1+k_1 \,r}{1 + k_2 \,r + k_3\, r^2}$ & $x \cdot \displaystyle \frac{1+k_1 \sqrt{1+c^2}\, x\, sgn(x)}{1+k_2 \,\sqrt{1+c^2}\, x \, sgn(x) + k_3 \,(1+c^2)\,x^2}$\\[3.5ex]
9 & $\displaystyle \frac{1+k_1\, r^2}{1 + k_2 \,r + k_3\, r^2}$ & $x \cdot \displaystyle \frac{1+k_1 \,(1+c^2)\,x^2}{1+k_2 \,\sqrt{1+c^2}\, x\, sgn(x) + k_3\, (1+c^2)\,x^2}$\\[3ex]\hline
\end {tabular}}}
\end{table*}

To be a candidate for radial distortion model, the function must satisfy the following properties:
\begin{itemize}
\item[{\rm \bf 1)}] This function is radially symmetric around the image center $(u_0, v_0)$ and it is expressible in terms of radius $r$ only;
\item[{\rm \bf 2)}] This function is continuous, hence $F(r) = 0$ iff $r = 0$;
\item[{\rm \bf 3)}] The resultant approximation of $x_d$ is an odd function of $x$.
\end{itemize}
Based on the above criteria, a new class of radial distortion models (model $\# \, 4, 5, 6, 7, 8, 9$) are proposed and summarized in Table \ref{table: Models}, where the other four polynomial models (model $\# \, 0, 1, 2, 3$) are also listed. 

Now, we want to compare the performance of our new radial distortion models with the four polynomial models in Table {\ref{table: Models}} based on the final value of objective function after nonlinear optimization by the Matlab function {\tt fminunc}. Using the public domain test images \cite{zhang98calibrationwebpage}, the desktop camera images (a color camera in our CSOIS), and the ODIS camera images \cite{LiliISIC03Flex} (the camera on ODIS robot built in our CSOIS), the final objective function $J$, the estimated distortion coefficients $\bf k$, and the 5 estimated intrinsic parameters ($\alpha, \beta, \gamma, u_0, v_0$) are shown in Tables \ref{table: Microsoft}, \ref{table: Desktop}, and \ref{table: ODIS}, respectively. The reason for listing ($\alpha, \beta, \gamma, u_0, v_0$) is to show that the estimated parameters after nonlinear optimization are consistent when using different distortion models. 

Table \ref{table: Microsoft}, \ref{table: Desktop}, and \ref{table: ODIS} are of the same format. The first column is the model number used in Table \ref{table: Models}. The second column shows the values of objective function $J$ defined in (\ref{eqn: objective function}). The third column, the rank, sorts the distortion models by $J$ in a [0-smallest, 9-largest] manner. 

After carefully examining Table \ref{table: Microsoft}, \ref{table: Desktop}, and \ref{table: ODIS}, we have the following observations:
\begin{itemize}

\item [\rm \bf 1)] Using the proposed rational models, we can achieve comparable, or even better, results compared with the polynomial models in Table \ref{table: Models}, where the best result is from model$_0$, model$_8$ or model$_9$. The advantage of using the last two models is that the inverse function is at most of order 3. The radial undistortion can thus be performed using the procedures described in Sec. \ref{sec: undistortion procedures};

\item [\rm \bf 2)] For each category of models, either polynomial or rational, they generally follow the trend that the more complex the model, the more accurate the performance (the smaller the objective function $J$);

\item [\rm \bf 3)] There is no general rule to decide at which point the polynomial models become better than the rational ones. It is dependent on the particular data set. However, the last three rational models always give the best results among all the non-iterative models, model 1-9;

\item [\rm \bf 4)] When the distortion is significant, the performance improvement using complex models is more obvious.

\end{itemize}

To make the results in this paper repeatable by other researchers for further investigation, we present the options we use for the nonlinear optimization: \texttt{options = optimset(`Display', `iter', `LargeScale', `off', `MaxFunEvals', 8000, `TolX', $10^{-5}$,  `TolFun', $10^{-5}$, `MaxIter', 120)}. The raw data of the extracted feature locations in the image plane are also available upon request. 


\begin{table*}[htb]
\centering
\caption{Comparison Results using Microsoft Images}
\label{table: Microsoft}
\renewcommand{\arraystretch}{1}
{\small
{\begin {tabular}{|c|c|c|rrr|rrrrr|}\hline
{$\bf \#$} & $\bf J$ & {\bf Rank} & \multicolumn{3}{|c|}{\bf Final Values of $k$} & \multicolumn{5}{|c|}{\bf Final Values of $(\alpha, \gamma, u_0, \beta, v_0)$} \\\hline
0 & 144.8802&{\bf 2}&-0.2286 &0.1905&  -       & 832.4860 &   0.2042 & 303.9605 & 832.5157 & 206.5811\\\hline
1 & 180.5714&  8&-0.0984 &        - &        - & 845.3051 &   0.1918 & 303.5723 & 845.2628 & 208.4394\\\hline      
2 & 148.2789&  7&-0.1984 &        - &        - & 830.7425 &   0.2166 & 303.9486 & 830.7983 & 206.5574\\\hline      
3 & 145.6592&  5&-0.0215 &  -0.1566 &        - & 833.6508 &   0.2075 & 303.9847 & 833.6866 & 206.5553\\\hline\hline 
4 & 185.0628&  9& 0.1031 &        - &        - & 846.1300 &   0.1921 & 303.5070 & 846.0823 & 208.6944\\\hline      
5 & 147.0000&  6& 0.2050 &        - &        - & 831.0863 &   0.2139 & 303.9647 & 831.1368 & 206.5175\\\hline      
6 & 145.4682&  4&-0.0174 &   0.1702 &        - & 833.3970 &   0.2071 & 303.9689 & 833.4324 & 206.5567\\\hline      
7 & 145.4504&{\bf 3}& 0.0170 &   0.1725 &        - & 833.3849 &   0.2068 & 303.9719 & 833.4198 & 206.5443\\\hline      
8 & 144.8328&{\bf 1}& 1.6457 &   1.6115 &   0.4054 & 830.9411 &   0.2044 & 303.9571 & 830.9705 & 206.5833\\\hline      
9 & 144.8257&{\bf 0}& 1.2790 &  -0.0119 &   1.5478 & 831.7373 &   0.2045 & 303.9573 & 831.7665 & 206.5925\\\hline
\end {tabular}}}
\end{table*}

\begin{table*}[htb]
\centering
\caption{Comparison Results using Desktop Images}
\label{table: Desktop}
\renewcommand{\arraystretch}{1}
{\small
{\begin {tabular}{|c|c|c|rrr|rrrrr|}\hline
{$\bf \#$} & $\bf J$ ($\times \, 10^3$) & {\bf Rank} & \multicolumn{3}{|c|}{\bf Final Values of $k$} & \multicolumn{5}{|c|}{\bf Final Values of $(\alpha, \gamma, u_0, \beta, v_0)$} \\\hline
0 & 0.7790 &{\bf 0} & -0.3435  &  0.1232  &       - & 277.1449 &  -0.5731 & 153.9882 & 270.5582 & 119.8105 \\\hline      
1 & 1.0167 & 8      & -0.2466  &       -  &       - & 295.5734 &  -0.8196 & 156.6108 & 288.8763 & 119.8528 \\\hline      
2 & 0.9047 & 7      & -0.2765  &       -  &       - & 275.5953 &  -0.6665 & 158.2016 & 269.2301 & 121.5257 \\\hline      
3 & 0.8033 & 6      & -0.1067  & -0.1577  &       - & 282.5642 &  -0.6199 & 154.4913 & 275.9019 & 120.0924 \\\hline\hline
4 & 1.2018 & 9      &  0.3045  &       -  &       - & 302.2339 &  -1.0236 & 160.5601 & 295.6767 & 120.7448 \\\hline      
5 & 0.7986 & 5      &  0.3252  &       -  &       - & 276.2521 &  -0.5780 & 154.7976 & 269.7064 & 120.3235 \\\hline      
6 & 0.7876 & 4      & -0.0485  &  0.2644  &       - & 279.5062 &  -0.5888 & 154.1735 & 272.8822 & 119.9564 \\\hline      
7 & 0.7864 &{\bf 3} &  0.0424  &  0.2834  &       - & 279.3268 &  -0.5870 & 154.1168 & 272.7049 & 119.9214 \\\hline      
8 & 0.7809 &{\bf 2} &  0.5868  &  0.5271  &  0.5302 & 275.8311 &  -0.5735 & 153.9991 & 269.2828 & 119.8195 \\\hline      
9 & 0.7800 &{\bf 1} &  0.2768  & -0.0252  &  0.6778 & 276.4501 &  -0.5731 & 153.9914 & 269.8850 & 119.8091 \\\hline      
\end {tabular}}}
\end{table*}

\begin{table*}[htb]
\centering
\caption{Comparison Results using ODIS Images}
\label{table: ODIS}
\renewcommand{\arraystretch}{1}
{\small
{\begin {tabular}{|c|c|c|rrr|rrrrr|}\hline
{$\bf \#$} & $\bf J$ ($\times \, 10^3$)& {\bf Rank} & \multicolumn{3}{|c|}{\bf Final Values of $k$} & \multicolumn{5}{|c|}{\bf Final Values of $(\alpha, \gamma, u_0, \beta, v_0)$} \\\hline
0  & 0.8403 &{\bf 2}&-0.3554  &  0.1633  &       - &  260.7658 &  -0.2741 & 140.0581 & 255.1489 & 113.1727\\\hline         
1  & 0.9444 &8      &-0.2327  &       -  &       - &  274.2660 &  -0.1153 & 140.3620 & 268.3070 & 114.3916\\\hline           
2  & 0.9331 &7      &-0.2752  &       -  &       - &  258.3193 &  -0.5165 & 137.2150 & 252.6856 & 115.9302\\\hline         
3  & 0.8513 &5      &-0.1192  & -0.1365  &       - &  266.0850 &  -0.3677 & 139.9198 & 260.3133 & 113.2412\\\hline\hline   
4  & 1.0366 &9      & 0.2828  &       -  &       - &  278.0218 &  -0.0289 & 139.5948 & 271.9274 & 116.2992\\\hline         
5  & 0.8676 &6      & 0.3190  &       -  &       - &  259.4947 &  -0.4301 & 139.1252 & 253.8698 & 113.9611\\\hline         
6  & 0.8450 &4      &-0.0815  &  0.2119  &       - &  264.4038 &  -0.3505 & 140.0528 & 258.6809 & 113.1445\\\hline         
7  & 0.8438 &{\bf 3}& 0.0725  &  0.2419  &       - &  264.1341 &  -0.3429 & 140.1092 & 258.4206 & 113.1129\\\hline         
8  & 0.8379 &{\bf 0}& 1.2859  &  1.1839  &  0.7187 &  259.2880 &  -0.2824 & 140.2936 & 253.7043 & 113.0078\\\hline         
9  & 0.8383 &{\bf 1}& 0.4494  & -0.0124  &  0.8540 &  260.9370 &  -0.2804 & 140.2437 & 255.3178 & 113.0561\\\hline         
\end {tabular}}}
\end{table*}

\section{Concluding Remarks}
\label{sec: conclusion}

This paper proposes a new class of rational radial distortion models. The appealing part of these distortion models is that they preserve high accuracy together with easy analytical undistortion formulae. Performance comparisons are made between this class of new rational models and the existing polynomial models. Experiments results are presented to show that this new class of rational distortion models can be quite accurate and efficient especially when the actual distortion is significant. 

\bibliography{calibration,csois1,csois2}
\end{document}